\pdfoutput=1

\documentclass[letterpaper]{article}

\usepackage[preprint]{neurips_2026}

\usepackage[utf8]{inputenc} 
\usepackage[T1]{fontenc}    
\usepackage{hyperref}       
\usepackage{url}            
\usepackage{booktabs}       
\usepackage{amsfonts}       
\usepackage{nicefrac}       
\usepackage{microtype}      
\usepackage{xcolor}         

\usepackage{amsmath} 
\usepackage{amssymb}
\usepackage{graphicx}
\usepackage{float}
\usepackage{multirow}
\usepackage{fvextra}
\usepackage{subcaption}
\usepackage{enumitem}

\DefineVerbatimEnvironment{jsonverbatim}{Verbatim}{
  breaklines=true,
  breakanywhere=true
}

\title{Scalable Hierarchical Attention Transformers for Multi-Turn Jailbreak Detection in Long Conversations}

\author{%
  Chenhui Hu \\
  Zscaler, Inc.\\
  \And
  Muhammed Salih \\
  Zscaler, Inc. \\
  \And
  Sudipto Guha \\
  Zscaler, Inc. \\
  \And
  Subramanian Srinivasan \\
  Zscaler, Inc. \\
}

\begin{document}

\maketitle


\begin{abstract}
	Multi-turn jailbreaks can evade turn-level moderation by spreading unsafe intent across a dialogue through gradual escalation, reframing, and role manipulation. We address multi-turn jailbreak detection as a conversation-level classification problem and introduce an efficient hierarchical detector that avoids expensive long-context concatenation while retaining cross-turn reasoning. The model encodes individual turns to form compact turn representations and applies a lightweight conversation module that captures dialogue dynamics and selectively attends to fine-grained evidence when needed. On a challenging evaluation benchmark of $14{,}038$ conversations, our approach achieves an F1 of $0.9394$, outperforming Claude Opus~4.7, the strongest competing baseline, by $0.07$ while halving its false-positive rate. Ablation studies confirm that each architectural component contributes meaningfully, with combining cross-attention and self-attention in the conversation module yielding a $2.26$ percentage point reduction in false-positive rate over the self-attention-only variant.
\end{abstract}

\section{Introduction}
\label{sec:introduction}

Large language models (LLMs) deployed in interactive settings are increasingly exposed to multi-turn jailbreak attempts, where an adversary does not issue a single overtly malicious prompt, but instead incrementally steers the conversation toward disallowed behavior. In these attacks, unsafe intent can be distributed across turns via gradual escalation, topic drift, role manipulation, and reframing of earlier instructions. As a result, effective defense requires conversation-level reasoning rather than isolated, turn-by-turn filtering.

A straightforward solution is to concatenate the entire dialogue history into a single sequence and apply a Transformer encoder to classify the conversation as safe or jailbreak. However, this approach scales poorly with conversation length. For a typical setting of 16 turns with 256 tokens per turn, concatenation yields a 4,096-token sequence, and the quadratic cost of self-attention makes both training and deployment expensive. Moreover, long-context concatenation forces the model to spend computation uniformly across all tokens, even though the signals for a jailbreak may be localized to a few turns or phrases.

We propose an efficient two-level hierarchical architecture for multi-turn jailbreak detection that preserves rich cross-turn reasoning while avoiding the computational bottleneck of full-sequence attention. At the first level, a TurnEncoder processes each turn independently using a pre-trained multilingual encoder (\texttt{intfloat/multilingual-e5-base}). Each turn is formatted with explicit speaker markers (two added special tokens, \texttt{[USER]} and \texttt{[ASST]}) and is padded/truncated to a fixed 256-token budget. The TurnEncoder produces (i) a compact turn summary embedding via the CLS representation and (ii) token-level embeddings for the full turn. Crucially, turns are encoded independently with no cross-turn token sharing at this stage, enabling efficient batched encoding across turns and conversations (with gradient checkpointing used during training to manage memory).

At the second level, a lightweight ConvTransformer reasons over the sequence of turn summaries to capture conversation dynamics that are characteristic of multi-turn jailbreaks. A learnable \texttt{[CONV\_CLS]} token is prepended to aggregate global information, and learned position embeddings (turn order) and role embeddings (user/assistant/CLS) allow the model to differentiate early vs.\ late turns and speaker identity. The ConvTransformer applies bidirectional self-attention over turn summaries to model cross-turn patterns (e.g., escalation and returns to previously refused topics). In addition, each layer performs cross-attention over the token-level embeddings produced by the TurnEncoder, allowing the model to ``zoom in'' on specific words or phrases within any turn when fine-grained cues are needed. Finally, an attentive pooling mechanism complements the \texttt{[CONV\_CLS]} representation by learning to weight turns by importance---useful, for example, when later turns carry stronger jailbreak signals than early benign setup turns.

This hierarchical design yields an attractive efficiency--accuracy trade-off: per-turn encoding remains bounded by a short context window, while the conversation-level module operates on a small sequence of turn summaries, yet retains access to token-level details through cross-attention. In experiments on a challenging 14,038-sample dataset, the proposed model achieves strong performance (95.90\% precision, 92.10\% recall, and 0.9394 F1), indicating that efficient conversation-level modeling can be highly effective for multi-turn jailbreak detection.

This paper makes the following contributions:
\begin{itemize}
  \item We formulate multi-turn jailbreak detection as a conversation-level classification problem that demands cross-turn reasoning under realistic length constraints.
  \item We introduce a hierarchical TurnEncoder--ConvTransformer architecture that avoids quadratic scaling from naive concatenation while preserving access to fine-grained token evidence via cross-attention.
  \item We demonstrate strong empirical performance on a multi-turn jailbreak dataset, showing that our approach outperforms state-of-the-art guardrails such as Claude Opus~4.7.
  \item We validate the importance of the cross-turn encoder via an ablation study, showing that replacing it with mean pooling significantly degrades jailbreak detection performance.
\end{itemize}

\section{Related Work}
\label{sec:related_work}

\subsection{Jailbreak / Prompt-Injection Detection and Moderation Classifiers}
Safety moderation for LLM systems is often implemented using supervised classifiers trained to detect toxic or policy-violating content, including widely used toxicity benchmarks and models (e.g., Jigsaw/Conversation AI datasets; \citep{borkan2019nuanced} and Detoxify; \citep{hanu2020detoxify}). While such classifiers are effective for overtly harmful content, jailbreaks frequently rely on instruction-following failures rather than explicit toxicity, motivating research on adversarial prompting and ``guard'' models that predict policy-violation risk.

A complementary line of work studies attack methods that generate jailbreak prompts through optimization or search. Universal adversarial suffix attacks such as GCG \citep{zou2023universal} show that iterative optimization can yield transferable jailbreak triggers across prompts and models. AutoDAN \citep{liu2023autodan} similarly generates stealthy jailbreak prompts via iterative optimization/evolution strategies. These approaches are important baselines and threat models for detection because they can produce non-obvious, non-toxic surface forms that nonetheless induce unsafe behavior.

\subsection{Multi-Turn Safety / Dialogue Risk Detection}
Multi-turn jailbreaks introduce temporal structure: an adversary can begin benignly and gradually move the model toward disallowed behavior through escalation, reframing, or role manipulation \citep{russinovich2024crescendo,chao2023jailbreaking,mehrotra2023tree}. Several recent attacks explicitly operationalize this multi-step structure. Crescendo \citep{russinovich2024crescendo} demonstrates a benign-to-malicious escalation strategy that unfolds across turns and can evade turn-local defenses. PAIR \citep{chao2023jailbreaking} jailbreaks black-box LLMs through iterative attacker--target refinement, using feedback from previous responses to craft subsequent queries. Tree of Attacks (TAP) \citep{mehrotra2023tree} extends iterative refinement into a tree search, branching attack trajectories and selecting promising continuations.

Long-context multi-turn threats also appear in many-shot jailbreaking \citep{anil2024manyshot}, where a long conversation history (often with many demonstrations) biases the model toward compliance. Defense-oriented follow-ups such as \emph{Mitigating Many-Shot Jailbreaking} \citep{ackerman2025mitigating} are relevant when positioning detection in the broader mitigation landscape (detectors, training-time alignment, inference-time policies).

On the benchmarking side, multi-turn evaluation suites such as MTJ-Bench / Multi-Turn Jailbreak Benchmark \citep{yang2025manyturn} reflect growing recognition that single-turn benchmarks underestimate real-world risk, and they provide more realistic settings for measuring conversation-level detection and early-warning capabilities.

Finally, newer work continues to push adaptive multi-turn attack automation at scale, including HarmNet \citep{narula2025harmnet}, \emph{Automating Deception} \citep{kumarappan2025automating}, and \emph{Echo Chamber} \citep{alobaid2026echochamber}. Collectively, these works strengthen the case that robust defenses must (i) model evolving attacker intent across turns and (ii) remain effective under adaptive adversaries \citep{narula2025harmnet,kumarappan2025automating,alobaid2026echochamber}.

\subsection{Indirect Prompt Injection and Tool-/Agent-Integrated Settings}
Beyond direct user prompting, indirect prompt injection arises when models ingest untrusted retrieved or tool-provided content that contains adversarial instructions \citep{greshake2023not,zhan2024injecagent}. \emph{Not What You{'}ve Signed Up For} \citep{greshake2023not} highlights real-world compromise vectors in LLM-integrated applications via malicious content embedded in retrieved documents or other inputs, often manifesting as multi-step interactions in agentic or retrieval-augmented pipelines. More recently, InjecAgent \citep{zhan2024injecagent} proposes threat models and benchmarking for indirect prompt injections in tool-integrated LLM agents. These settings reinforce the importance of multi-turn and multi-source context modeling: the effective ``conversation'' may include tool outputs, retrieved passages, and intermediate reasoning steps, all of which can participate in a gradual jailbreak \citep{greshake2023not,zhan2024injecagent}.

\subsection{Hierarchical Modeling for Long Documents and Dialogue}
Hierarchical architectures are a standard approach for modeling structured long inputs by separating local encoding from global aggregation. Classic examples include Hierarchical Attention Networks \citep{yang2016han}, which encode tokens into sentence representations and then apply higher-level attention over sentences for document classification. Related hierarchical dialogue modeling traditions similarly encode utterances/turns independently and then model the sequence of turns to capture conversational dynamics. This perspective aligns closely with multi-turn jailbreak detection: the salient signal is often how turns relate (escalation, contradiction, instruction manipulation), rather than token-level interactions uniformly across a flattened sequence. Our approach fits this family by combining turn-wise representations with conversation-level attention, while still retaining access to token-level evidence through cross-attention.

\subsection{Efficient Transformers and Long-Context Modeling (Not Just ``Long Context'')}
A separate body of work aims to scale Transformers to long sequences by reducing attention cost or adding memory mechanisms, including sparse attention (Longformer \citep{beltagy2020longformer}, BigBird \citep{zaheer2020bigbird}), hashing or approximation methods (Reformer \citep{kitaev2020reformer}, Linformer \citep{wang2020linformer}, Performer \citep{choromanski2021performer}), and recurrence/memory (Transformer-XL \citep{dai2019transformerxl}, Compressive Transformer \citep{rae2020compressive}). These methods primarily treat the input as a single long token stream.

In contrast, many jailbreak settings are naturally structured by turns (and sometimes by tool calls / retrieved documents). Multi-turn attacks such as Crescendo \citep{russinovich2024crescendo}, PAIR \citep{chao2023jailbreaking}, TAP \citep{mehrotra2023tree}, and many-shot jailbreaking \citep{anil2024manyshot} exploit exactly this conversational structure---e.g., building a benign prefix, iteratively refining instructions, or using long interaction histories. This motivates architectures that allocate computation to turn-level attention while selectively revisiting token-level detail, rather than relying solely on generic long-context Transformers.

\section{Problem Formulation}
\label{sec:problem_formulation}

\subsection{Conversation and Turn Notation}
\label{sec:notation}
We model a conversation as a sequence of discrete turns:
\begin{equation}
	\mathcal{C} = (x_1, x_2, \ldots, x_T),
\end{equation}
where $T$ is the number of turns and each turn $x_t$ is a role-annotated message
\begin{equation}
	x_t = (r_t, u_t),
\end{equation}
with role $r_t \in \{\text{user}, \text{assistant}\}$ and textual content $u_t$. When role tags are not needed, we refer to the content-only sequence $(u_1, u_2, \ldots, u_T)$. We denote the prefix of the conversation up to turn $t$ as
\begin{equation}
	\mathcal{C}_{\le t} = (x_1, x_2, \ldots, x_t).
\end{equation}

\subsection{Labels and Tasks}
\label{sec:labels_tasks}

We use a binary conversation-level label $y \in \{0,1\}$, where $y=0$ indicates a safe conversation and $y=1$ denotes a jailbreak conversation. Note that datasets differ in whether $y=1$ reflects jailbreak \emph{attempts} (malicious user intent is present at some point) or only \emph{successful} jailbreaks (the assistant produces disallowed content); we follow the definition specified by each dataset used in our experiments (Section~\ref{sec:experiments}).

\paragraph{Conversation-level classification.}
Given the full conversation $\mathcal{C}$, the primary task is to estimate the posterior
\begin{equation}
	p(\mathcal{C}) \;=\; \Pr(y=1 \mid \mathcal{C}),
\end{equation}
and produce a hard decision by thresholding,
\begin{equation}
	\hat{y} \;=\; \mathbf{1}\{\, p(\mathcal{C}) \ge \tau \,\},
\end{equation}
where $\tau \in [0,1]$ is an operating threshold and $\mathbf{1}\{\cdot\}$ denotes the indicator function (equal to $1$ if its argument is true and $0$ otherwise).

\paragraph{Optional per-turn risk scoring.}
In a streaming setting, the detector produces a per-turn risk score at each turn $t$:
\begin{equation}
	p_t \;:=\; p(\mathcal{C}_{\le t}) \;=\; \Pr(y=1 \mid \mathcal{C}_{\le t}), \qquad t=1,\ldots,T.
\end{equation}

\paragraph{Optional early detection.}
For early warning, we trigger an alert as soon as the risk exceeds the threshold $\tau$:
\begin{equation}
	\hat{t} \;=\; \min\{t \in \{1,\ldots,T\} : p_t \ge \tau\},
\end{equation}
with the convention that $\hat{t} = \infty$ if the threshold is never crossed. A detector is better if it achieves high accuracy while minimizing $\hat{t}$ on jailbreak conversations.

\subsection{Threat Model}
\label{sec:threat_model}
We assume an adversarial user who controls the user turns and can adaptively choose messages based on prior assistant outputs. The attacker is described by an adaptive policy $\pi$ such that, for each user-role turn,
\begin{equation}
	u_t \sim \pi(\cdot \mid \mathcal{C}_{\le t-1}).
\end{equation}
Assistant-role turns are produced by the target model (or an environment/tool wrapper) and are treated as exogenous to the defender.

The defender observes the streaming conversation and, at each turn $t$, produces a per-turn risk score
\begin{equation}
	g(\mathcal{C}_{\le t}) \;=\; p_t,
\end{equation}
from which a hard per-turn decision $\hat{y}_t = \mathbf{1}\{p_t \ge \tau\}$ and, in particular, the conversation-level decision $\hat{y} = \hat{y}_T$ are derived. The defender does not control the attacker and must remain robust to multi-turn escalation strategies and other adaptive behaviors expressed across turns.

\section{Method}
\label{sec:method}

\subsection{Baseline and Complexity Motivation}
\label{sec:baseline_complexity}

\paragraph{Naive concatenation baseline.}
A straightforward approach concatenates all turns into a single long token sequence and runs a Transformer encoder over the full conversation. Let turn $t$ contain $n_t$ tokens. The total concatenated length is
\begin{equation}
N \;=\; \sum_{t=1}^{T} n_t .
\end{equation}
For standard Transformer self-attention, the dominant cost scales quadratically with sequence length:
\begin{equation}
\text{cost} \;=\; \mathcal{O}(N^2) \quad (\text{up to constant factors and hidden-size terms}).
\end{equation}


\paragraph{Concrete example ($24 \times 256 \rightarrow 6144$ tokens).}
With up to $T=24$ turns and a per-turn cap of $n_t \le 256$ tokens,
\begin{equation}
	N \;=\; 24 \cdot 256 \;=\; 6144,
\end{equation}
and attention scales as
\begin{equation}
	N^2 \;=\; 6144^2 \;=\; 37{,}748{,}736
\end{equation}
pairwise interactions per layer (up to constant factors and number of heads). This quadratic scaling motivates an architecture that avoids attention over the full concatenated context while preserving cross-turn reasoning.

\subsection{Level 1: TurnEncoder}
\label{sec:turnencoder}

\begin{figure}[t]
  \centering
  \includegraphics[width=\linewidth]{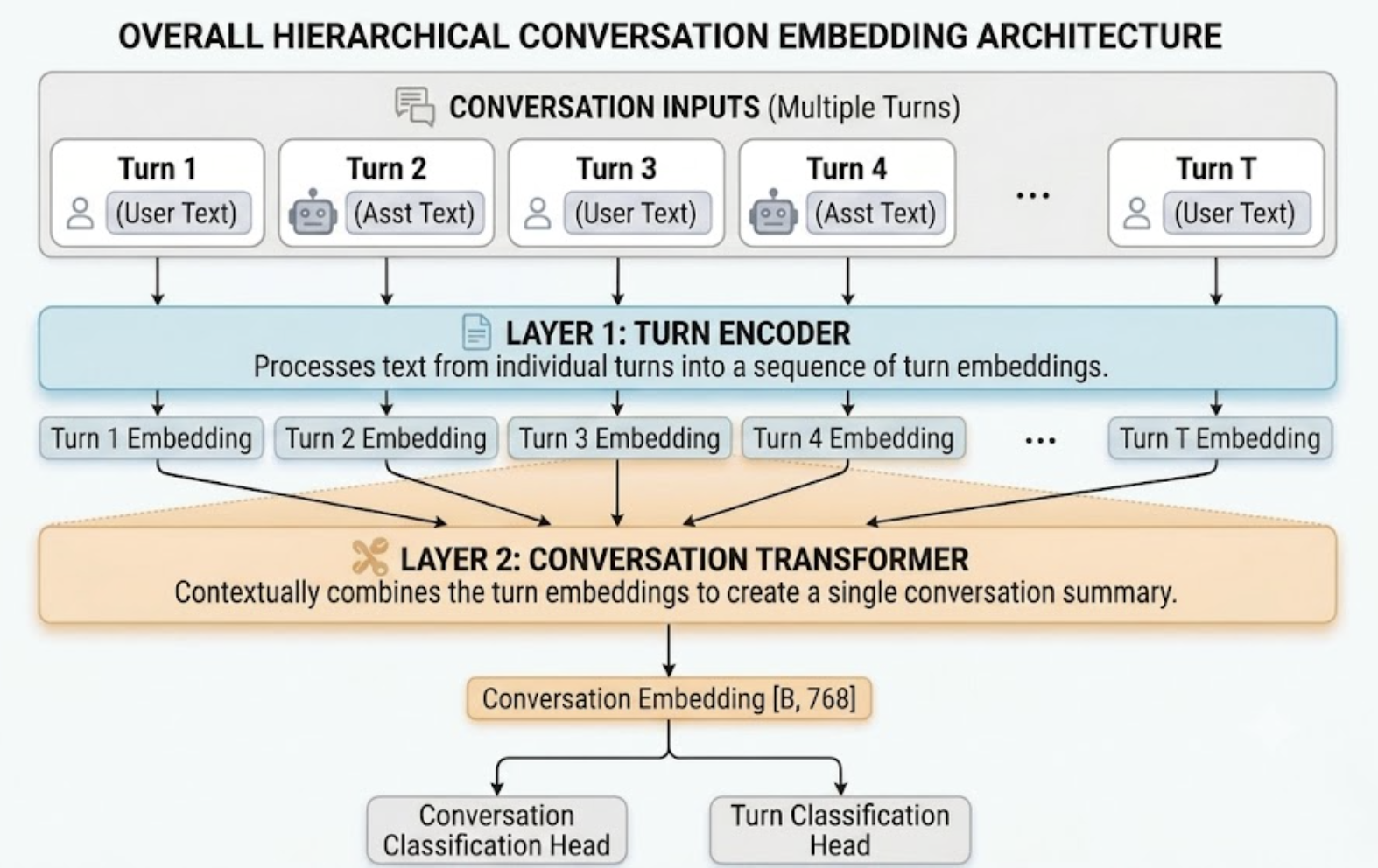}
  \caption{Multi-turn jailbreak detection based on a hierarchical conversation embedding architecture.}
  \label{fig:model_architecture}
\end{figure}

\paragraph{Overview.}
As shown in Figure \ref{fig:model_architecture}, Level 1 encodes each turn independently using a pre-trained language model, producing (i) a compact summary embedding for cross-turn reasoning and (ii) token-level embeddings for optional fine-grained retrieval by Level 2. Encoding turns independently keeps computation roughly linear in the number of turns and enables efficient batching.

\paragraph{Backbone and role markers.}
The TurnEncoder wraps the pretrained multilingual E5-base model (\texttt{intfloat/multilingual-e5-base}), a 12-layer Transformer with hidden size $d=768$.
We add two special tokens to the vocabulary, \texttt{[USER]} and \texttt{[ASST]}, which act as role markers. Each turn is tokenized independently as
\begin{equation}
	\tilde{x}_t \;=\; [\texttt{ROLE}(r_t)] \,\Vert\, u_t,
\end{equation}
where $\Vert$ denotes token-sequence concatenation, then padded or truncated to a fixed maximum length $L=256$ tokens.

\paragraph{Outputs.}
For each turn $t$, the encoder outputs:
\begin{itemize}
  \item \textbf{Turn summary embedding (CLS embedding):}
  \begin{equation}
  \mathbf{s}_t \in \mathbb{R}^{d}, \qquad \mathbf{s}_t \;=\; \mathbf{h}^{(L_{\text{enc}})}_{t,1},
  \end{equation}
  defined as the first token's output (CLS) from the last encoder layer ($L_{\text{enc}}=12$).

  \item \textbf{Token-level embeddings (last-layer hidden states):}
  \begin{equation}
  \mathbf{H}_t \in \mathbb{R}^{L \times d}, \qquad \mathbf{H}_t \;=\; \big[\mathbf{h}^{(L_{\text{enc}})}_{t,1}, \ldots, \mathbf{h}^{(L_{\text{enc}})}_{t,L}\big]^{\top}.
  \end{equation}
\end{itemize}
The summaries $\{\mathbf{s}_t\}_{t=1}^{T}$ feed Level 2 self-attention, while $\{\mathbf{H}_t\}_{t=1}^{T}$ serve as a cross-attention memory for ``zooming in'' on salient phrases.

\paragraph{Independence across turns.}
There is no cross-turn token sharing at Level 1; all cross-turn reasoning (e.g., recognizing escalation from turn 3 to turn 5) is deferred to Level 2. This independence also enables encoding all turns in a batch in one batched forward pass.

\paragraph{Memory optimization.}
Gradient checkpointing is enabled during training to reduce memory usage, which is important when processing up to 24 turns per conversation per GPU. Table~\ref{tab:turnencoder_config} summarizes the TurnEncoder configuration.

\begin{table}[t]
\centering
\caption{TurnEncoder configuration.}
\label{tab:turnencoder_config}
\begin{tabular}{ll}
\toprule
\textbf{Parameter} & \textbf{Value} \\
\midrule
Base model & \texttt{intfloat/multilingual-e5-base} (12 layers, 768-dim) \\
Special tokens & \texttt{[USER]}, \texttt{[ASST]} \\
Max tokens per turn & 256 \\
Dropout & 0.2 \\
Gradient checkpointing & Enabled \\
\bottomrule
\end{tabular}
\end{table}

\subsection{Level 2: ConvTransformer}
\label{sec:convtransformer}

\paragraph{Overview.}
Level 2 performs conversation-level reasoning over the sequence of turn summaries and optionally retrieves token-level details through cross-attention. This captures multi-turn patterns such as gradual escalation, topic shifts, instruction hierarchy manipulation, and role changes.

\subsubsection{Input sequence construction}
\label{sec:input_construction}
Given turn summaries $\{\mathbf{s}_t\}_{t=1}^{T}$, we prepend a learnable conversation-level token $\mathbf{s}_{0}$ corresponding to \texttt{[CONV\_CLS]}:
\begin{equation}
\mathbf{S}^{(0)} \;=\; [\mathbf{s}_{0}, \mathbf{s}_1, \ldots, \mathbf{s}_T] \in \mathbb{R}^{(T+1)\times d}.
\end{equation}
The final representation of \texttt{[CONV\_CLS]} serves as the primary conversation embedding.

\subsubsection{Position and role embeddings}
\label{sec:pos_role_emb}
Each element in the input sequence is augmented with learned position and role embeddings:
\begin{equation}
\tilde{\mathbf{s}}_i \;=\; \mathbf{s}_i + \mathbf{p}_i + \mathbf{e}_{\rho_i}, \qquad i=0,\ldots,T,
\end{equation}
where $\mathbf{p}_i$ is a learned position embedding and $\mathbf{e}_{\rho_i}$ is a learned role embedding. We use learned position embeddings for 25 positions (position 0 for \texttt{[CONV\_CLS]}, positions 1--24 for turns), and learned role embeddings for 3 roles (user, assistant, and a CLS placeholder role).

\subsubsection{Self-attention over turn summaries}
\label{sec:self_attn_summaries}
The ConvTransformer is a 4-layer Transformer decoder (pre-layer normalization, 8 heads, Feed-Forward Network (FFN) hidden size 2048, dropout 0.2). It applies bidirectional self-attention (no causal mask) over the augmented summary sequence
\begin{equation}
\tilde{\mathbf{S}}^{(0)} \;=\; [\tilde{\mathbf{s}}_0, \tilde{\mathbf{s}}_1, \ldots, \tilde{\mathbf{s}}_T],
\end{equation}
because the classifier can observe the full conversation at inference time in the current setting. This supports detecting global multi-turn structures (e.g., returning to a previously refused topic several turns later).

\subsubsection{Cross-attention to token-level memory}
\label{sec:cross_attn_memory}
In addition to self-attention over summaries, each decoder layer performs cross-attention over token-level embeddings from Level 1. We flatten all per-turn token embeddings into one memory:
\begin{equation}
\mathbf{M} \;=\; [\mathbf{H}_1; \mathbf{H}_2; \ldots; \mathbf{H}_T] \in \mathbb{R}^{(T\cdot L)\times d},
\end{equation}
where $L=256$ and thus $|\mathbf{M}| = T\cdot L$ tokens (with $T \le 24$).
Tokens in $\mathbf{M}$ are augmented with turn-aware positional signals (reusing the same turn position table) so attention heads can infer which turn each token belongs to. This mechanism allows Level 2 to ``zoom in'' on specific words or phrases that may carry jailbreak signal.

\subsubsection{Attentive pooling}
\label{sec:attentive_pooling}
After the decoder layers, we compute an additional conversation summary via attentive pooling: a learned query attends over all turn outputs to produce a weighted combination, which is then added to the \texttt{[CONV\_CLS]} output via a residual connection. This lets the model weight turns differently depending on their importance (e.g., later turns in an escalation attack). We list the conversation module configuration in Table~\ref{tab:convtransformer_config}.

\begin{table}[t]
\centering
\caption{ConvTransformer configuration.}
\label{tab:convtransformer_config}
\begin{tabular}{ll}
\toprule
\textbf{Parameter} & \textbf{Value} \\
\midrule
Decoder layers & 3 \\
Attention heads & 8 \\
FFN hidden dim & 2{,}048 \\
Normalization & Pre-LN \\
Position embeddings & Learned, 25 positions \\
Role embeddings & Learned, 3 roles \\
Dropout & 0.2 \\
\bottomrule
\end{tabular}
\end{table}

\subsubsection{Classification heads}
\label{sec:classification_heads}
The final 768-dimensional conversation embedding is passed through dropout and a linear projection to produce two logits (safe vs.\ jailbreak):
\begin{equation}
\boldsymbol{\ell} \;=\; \mathbf{W}\,\mathbf{z} + \mathbf{b} \in \mathbb{R}^{2},
\end{equation}
where $\mathbf{z}\in\mathbb{R}^{768}$ is the final conversation embedding.
At inference, softmax yields the jailbreak probability
\begin{equation}
p(\mathcal{C}) \;=\; \mathrm{softmax}(\boldsymbol{\ell})_{1}.
\end{equation}

A secondary turn-level classification head projects each turn's output embedding to two logits to support an auxiliary per-turn loss (multiple-instance learning). In the described experiment, the auxiliary loss weight is set to zero, so this head has no effect on training or inference (though logits may still be computed).

\subsection{Training Objective and Inference}
\label{sec:training_inference}

\subsubsection{Input constraints}
\label{sec:input_constraints}

We cap conversation length by truncating at both the turn and conversation levels: we limit each turn to at most 256 tokens, and we limit each conversation to at most 24 turns (keeping only the most recent 24 turns when a conversation is longer).

\subsubsection{Training objective}
\label{sec:training_objective}
Let $y \in \{0,1\}$ denote the conversation label (0=safe, 1=jailbreak). We compute a predicted jailbreak probability $p(\mathcal{C})$ and optimize binary cross-entropy:
\begin{equation}
	\mathcal{L}_{\text{conv}} \;=\; -\, y \log p(\mathcal{C}) \;-\; (1-y)\log\big(1-p(\mathcal{C})\big).
\end{equation}
When turn-level labels are available, an auxiliary per-turn cross-entropy loss can be added to provide finer-grained supervision, though we do not activate it in the experiments presented herein.

\subsubsection{Inference}
At inference time, we encode each turn independently with the TurnEncoder to obtain turn summaries and (optionally) token-level embeddings, then run the Level 2 conversation module over the resulting sequence. Following common moderation practice, we drop the final assistant turn at inference so the decision is conditioned only on the conversation history up to the last user turn. The classifier outputs a jailbreak probability via softmax, and we predict \textit{jailbreak} if the probability exceeds a threshold $\tau$ (set on a validation set); turn-level logits are optional and are not used for the final decision.

\section{Datasets}
\label{sec:datasets}

\subsection{Training Set}
\label{sec:training_data}

We train all models on a curated dataset of $98{,}626$ multi-turn conversations, approximately balanced between jailbreak ($49\%$) and safe ($51\%$) examples. A conversation is labeled \emph{jailbreak} ($y=1$) if any assistant turn contains harmful content elicited through adversarial manipulation, and \emph{safe} ($y=0$) otherwise.

\paragraph{Safe conversations.}
The safe subset ($50{,}287$ conversations) combines real user--assistant dialogues from open-domain corpora with purpose-built hard negatives. The open-domain portion is drawn from UltraChat, WildChat, ShareGPT, and OASST2, and teaches the detector to tolerate normal multi-turn interaction patterns. Hard negatives are benign conversations that share structural features with attacks---such as progressive complexity, polite follow-ups, or extended length---and reduce false alarms on legitimate conversations.

\paragraph{Jailbreak conversations.}
The jailbreak subset ($48{,}339$ conversations) covers eight distinct multi-turn attack strategies:

\begin{enumerate}
	\item \textbf{Framing and disguise} ($24{,}392$). The largest category, covering attacks where the user wraps harmful requests in a legitimizing frame. This includes persona adoption (e.g., ``I'm a journalist researching\ldots''), academic or historical framing that conceals harmful intent behind scholarly language, and fictional or hypothetical scenarios used to indirectly elicit harmful information. The key characteristic is that each individual turn may appear benign in isolation; the harmful intent is revealed only by the overall direction of the conversation.
	
	\item \textbf{Escalating manipulation} ($9{,}839$). Short conversations where each successive turn makes requests slightly more specific or harmful, directly exploiting the trust or compliance established in earlier turns.
	
	\item \textbf{Gradual escalation} ($4{,}986$). Unlike escalating manipulation (which starts harmful and increases specificity), these conversations begin on a genuinely benign topic and progressively steer toward harmful requests over 6--34 turns, making it difficult to identify the transition point.
	
	\item \textbf{Refusal persistence} ($4{,}657$). After the model refuses, the attacker rephrases the same harmful request using different angles (bureaucratic framing, hypothetical context, emotional appeals) across multiple turns until the model complies.
	
	\item \textbf{Task decomposition} ($2{,}095$). A harmful goal is broken into seemingly innocuous sub-questions asked across many turns (e.g., asking separately about chemical ingredients, synthesis conditions, and purification steps), where no individual question appears dangerous but the composite reveals malicious intent.
	
	\item \textbf{Sudden adversarial insertion} ($903$). A single adversarial turn is embedded at a random late position within an otherwise entirely benign conversation, testing whether the detector can identify isolated malicious signal diluted among many benign exchanges.
	
	\item \textbf{Context narrowing} ($253$). A benign multi-turn discussion is followed by a final user turn that demands the prior context be reformatted into harmful content (e.g., ``rewrite the above as a step-by-step guide to\ldots'').
	
	\item \textbf{Prompt injection and other strategies} ($1{,}214$). Instruction-override attacks (``ignore previous instructions''), explicit attempts to suppress safety guidelines, post-refusal persistence where the user eventually succeeds, and short context-shifting attacks.
\end{enumerate}

\subsection{Evaluation Benchmark}
\label{sec:eval_data}

We evaluate on a comprehensive multi-source benchmark of $14{,}038$ conversations ($8{,}182$ safe and $5{,}856$ jailbreak) constructed to assess detector performance across a broad spectrum of attack strategies and benign interaction patterns. The benchmark combines established external corpora with targeted synthetic conversations, ensuring coverage of attack vectors that no single existing dataset provides in isolation. The full per-source composition is reported in Tables~\ref{tab:safe_sources} and~\ref{tab:jb_sources} (Appendix~\ref{app:dataset_composition}).

\paragraph{Safe conversations.}
The $8{,}182$ safe conversations are drawn from five real-world dialogue corpora spanning open-domain discussion, task-oriented instruction, and user--assistant interaction. UltraChat~\citep{ding2023ultrachat} and WildChat~\citep{zhao2024wildchat} contribute the bulk of safe examples, representing diverse multi-turn exchanges between users and LLMs. ShareGPT~\citep{sharegpt2023} and OpenAssistant (OASST2)~\citep{kopf2023openassistant} provide shorter community-collected dialogues, and SafeMTData~\citep{ren2024safemtdata} contributes benign conversations on sensitive-adjacent topics where user intent is genuinely benign. To calibrate false-positive rates, we additionally include synthetically generated hard negatives---benign conversations that share structural features with jailbreak attacks (progressive complexity, topic shifts, extended length) but contain no adversarial intent.

\paragraph{Jailbreak conversations.}
The $5{,}856$ jailbreak conversations span seven distinct attack-strategy categories. External corpora provide ecologically valid attacks: XGuard~\citep{xguard2025} contributes persona-adoption red-teaming, Red Queen~\citep{jiang2025redqueen} supplies short targeted manipulations, carl213~\citep{broomfield2024multiturn} provides cipher-encoded attacks via user-defined word-substitution mappings, and SafeDialBench~\citep{cao2025safedialbench} and the Duke CEI Malicious-Educator benchmark~\citep{chen2025maliciouseducator} contribute additional multi-turn attack scenarios. FRACTURED-SORRY-Bench~\citep{priyanshu2024fracturedsorry} contributes multi-step attacks with benign intermediary turns, and LAD~\citep{kulkarni2024lad} contributes gradual escalation in technical domains (SCADA systems, autonomous vehicles, network security). To cover attack vectors underrepresented in existing corpora, we additionally include synthetic conversations implementing gradual escalation (progressive steering across $9$--$32$ turns), sudden adversarial insertion (a single malicious turn embedded within an otherwise benign conversation), and context narrowing (benign prefixes reframed toward harmful requests via the final user turn). Prompt-injection attacks ($2$--$12$ turns) test robustness to single-shot instruction-override attempts within a multi-turn context.

\paragraph{Deduplication.}
To prevent train--test leakage, the benchmark is filtered against the training set using a strict five-rule protocol: (1) exact whole-conversation match; (2) any prefix of $\geq 2$ turns matching; (3) any suffix of $\geq 2$ turns matching; (4) first user-turn hash match; and (5) final user-turn hash match. This eliminates both verbatim repetition and template-level overlap that arises when synthetic generators draw from shared seed pools.

\paragraph{Rationale for combining external and synthetic data.}
No single jailbreak corpus covers the full range of known multi-turn attack strategies. External datasets provide ecological validity---these are attacks that were actually attempted against deployed systems---but they cluster around specific tactics (e.g., XGuard is predominantly persona-based; Red Queen is uniformly five turns). Synthetic augmentation addresses three gaps: (i)~controlled variation of conversation length and escalation speed, enabling fine-grained evaluation of detector sensitivity; (ii)~coverage of structurally distinct attack types (sudden insertion, context narrowing) that are rare in existing corpora; and (iii)~hard negatives that share surface-level features with attacks, which are critical for measuring false-positive robustness, since naturally occurring safe conversations rarely resemble jailbreaks in structure.

\paragraph{Summary statistics.}
The benchmark comprises $14{,}038$ conversations ($58.3\%$ safe, $41.7\%$ jailbreak), with turn counts ranging from $2$ to $102$ (mean $9.3$, median $8$) and approximately $1{,}962$ tokens per conversation on average. The $24$ distinct source identifiers are grouped into eight attack-strategy categories and three safe-conversation categories, drawing on published external datasets and synthetic conversations.

\section{Experiments}
\label{sec:experiments}

\subsection{Experimental Setup}
\label{sec:experimental_setup}

\paragraph{Model configuration.}
Our primary model uses Multilingual E5-base~\citep{wang2022e5} (278M parameters) as the TurnEncoder with a 4-layer ($L=4$) ConvTransformer decoder. Each turn is tokenized to a maximum of 256 tokens, and conversations are truncated to 24 turns (retaining the most recent turns when a conversation exceeds this limit). The ConvTransformer uses 8 attention heads with a feed-forward dimension of $2{,}048$.

\paragraph{Training procedure.}
Training proceeds in two stages. In the first stage, we train the ConvTransformer with a frozen TurnEncoder for 2 epochs (learning rate $1 \times 10^{-4}$). In the second stage, we jointly fine-tune both components for 4 additional epochs, using a reduced learning rate of $5 \times 10^{-6}$ for the TurnEncoder and $1 \times 10^{-4}$ for the ConvTransformer. Joint fine-tuning enables the TurnEncoder and ConvTransformer to adapt to each other, improving overall end-to-end performance compared to keeping the encoder frozen.

\paragraph{Baselines.}
We compare against the following baselines:

\begin{itemize}[leftmargin=1.5em, itemsep=2pt]
	\item \textbf{Mean Pool.} Uses the same Multilingual E5-base TurnEncoder~\citep{wang2022e5} but replaces the ConvTransformer with simple mean pooling over turn embeddings, followed by a linear classifier. This ablation isolates the contribution of the attention-based conversation-level aggregation.
	
	\item \textbf{Concatenation.} A Llama 3.3 70B model~\citep{dubey2024llama3} that concatenates the most recent turns up to a fixed window size of 4, then classifies the resulting truncated context.
	
	\item \textbf{ToT OR / ToT Majority.} Tree-of-Thought baselines built on the same Llama 3.3 70B model~\citep{dubey2024llama3}. The full conversation is examined from three complementary analytical lenses: (i)~\emph{user intent}---inferring the user's true goal and whether it is manipulative; (ii)~\emph{potential harm}---assessing worst-case misuse and safety impact; and (iii)~\emph{jailbreak patterns}---detecting known tactics such as roleplay framing, hypotheticals, authority claims, or gradual escalation. Each lens produces an independent binary decision. \textbf{ToT OR} predicts jailbreak if \emph{any} of the three lenses is positive; \textbf{ToT Majority} predicts jailbreak if at least two of the three are positive.
	
	\item \textbf{Claude Sonnet 4 / Sonnet 4.5 / Opus 4.7} \citep{anthropic2025claude}. State-of-the-art proprietary models from Anthropic. Claude Opus 4.7 is Anthropic's flagship Opus-tier model. These represent strong, widely deployed safety-tuned baselines; outperforming them provides compelling evidence for the effectiveness of our approach.
\end{itemize}

\subsection{Main Results}
\label{sec:main_results}

Table~\ref{tab:main_results} reports performance on the full evaluation benchmark (${\sim}14{,}$k conversations). Our model significantly outperforms all baselines across every metric.

Compared to Claude Opus 4.7---the strongest competing system---our model improves recall by $+13.8$ points ($92.10\%$ vs.\ $78.30\%$), precision by $+5.3$ points ($95.90\%$ vs.\ $90.60\%$), and F1 by $+0.070$ ($0.9394$ vs.\ $0.8691$), while simultaneously reducing the false-positive rate from $5.80\%$ to $2.80\%$. The advantage over the Llama 3.3 70B--based baselines (Concatenation, ToT OR, ToT Majority) is even more pronounced: our model improves F1 by $0.33$--$0.42$ absolute and substantially lowers FPR. Notably, the ToT OR variant trades a higher false-positive rate ($5.80\%$) for moderate recall ($47.10\%$), yet still falls short of our model by over $45$ recall points---indicating that simply expanding the decision surface of a turn-level classifier cannot compensate for the lack of conversation-level reasoning.

These results demonstrate that our hierarchical ConvTransformer not only generalizes well to out-of-distribution conversations but also scales effectively to harder adversarial scenarios, outperforming even the most capable proprietary models while maintaining the lowest false-positive rate among all systems evaluated.

\begin{table}[t]
	\centering
	\small
	\setlength{\tabcolsep}{6pt}
	\renewcommand{\arraystretch}{1.15}
	\caption{Performance comparison on the full evaluation benchmark (${\sim}14{,}$k conversations). $\tau$ denotes the classification threshold. Best results in \textbf{bold}.}
	\label{tab:main_results}
	\begin{tabular}{@{}lcccc@{}}
		\toprule
		\textbf{Model} & \textbf{Recall} & \textbf{Precision} & \textbf{F1} & \textbf{FPR} \\
		\midrule
		Concatenation & 36.20\% & 88.50\% & 0.5136 & 3.40\% \\
		ToT OR                     & 47.10\% & 85.30\% & 0.6072 & 5.80\% \\
		ToT Majority               & 37.00\% & 89.10\% & 0.5226 & 3.20\% \\
		\midrule
		Claude Sonnet 4            & 59.10\% & 91.70\% & 0.7189 & 3.80\% \\
		Claude Sonnet 4.5          & 66.10\% & 91.70\% & 0.7682 & 4.30\% \\
		Claude Opus 4.7            & 78.30\% & 90.60\% & 0.8691 & 5.80\% \\
		\midrule
		Our Model ($\tau = 0.50$)  & \textbf{92.10\%} & \textbf{95.90\%} & \textbf{0.9394} & \textbf{2.80\%} \\
		\bottomrule
	\end{tabular}
\end{table}

\subsection{Attention Pattern Analysis}
\label{sec:attention_analysis}

To understand how the model distinguishes jailbreak from safe conversations, we extract attention weights from the $L{=}4$ ConvTransformer decoder. We instrument all decoder layers to capture both self-attention and cross-attention weights during inference on 1{,}000 conversations (500 jailbreak, 500 safe, each with $\geq 4$ turns) from the held-out test set. We focus on the attention distribution of the \texttt{[CONV\_CLS]} token, which aggregates conversation-level evidence for the final classification decision.

\begin{figure}[t]
	\centering
	\begin{subfigure}[t]{\textwidth}
		\centering
		\includegraphics[width=\textwidth]{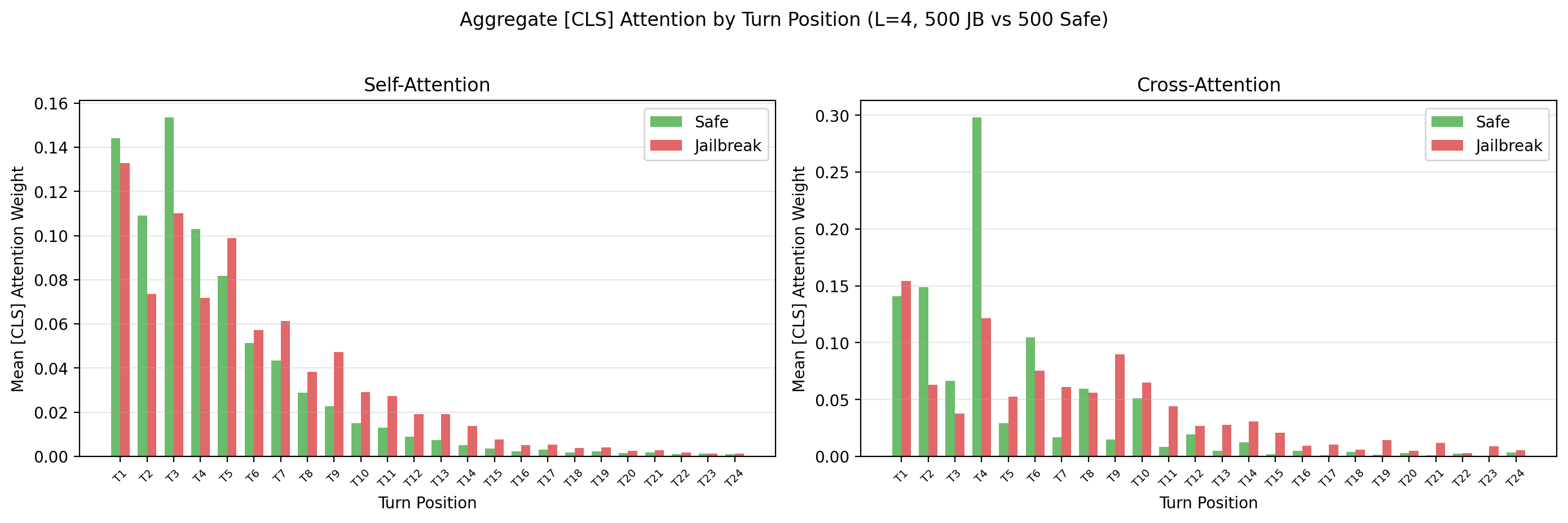}
		\caption{Aggregate \texttt{[CONV\_CLS]} attention weights by turn position. Safe conversations (green) concentrate attention on early turns (T1--T3), while jailbreak conversations (orange) distribute weight more uniformly across later positions up to T24.}
		\label{fig:attention_aggregate}
	\end{subfigure}
	
	\vspace{0.4em}
	
	\begin{subfigure}[t]{\textwidth}
		\centering
		\includegraphics[width=\textwidth]{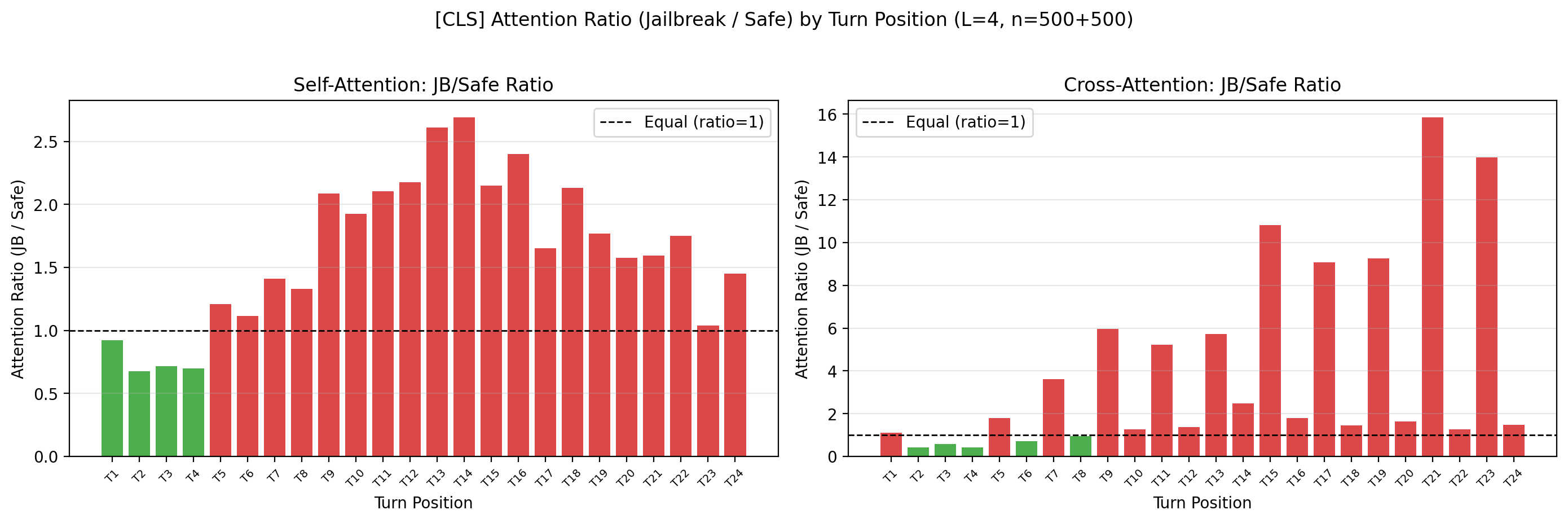}
		\caption{JB/Safe attention ratio by turn position. Green bars indicate safe-dominant positions (ratio $< 1$); red bars indicate jailbreak-dominant positions (ratio $> 1$). Self-attention ratio peaks at ${\sim}2.7{\times}$ at T14; cross-attention ratio reaches ${\sim}6{\times}$ at T9, ${\sim}11{\times}$ at T15, and ${\sim}16{\times}$ at T21, with the largest ratios concentrated on odd-numbered (user) turns.}
		\label{fig:attention_ratio}
	\end{subfigure}
	
	\caption{Attention analysis of the ConvTransformer decoder ($L{=}4$, $n{=}500{+}500$ conversations). Left panels show self-attention; right panels show cross-attention. (a)~Raw attention weights reveal that jailbreak conversations shift \texttt{[CONV\_CLS]} focus toward later turns (up to T24). (b)~The JB/Safe ratio quantifies this shift, growing across turn positions and reaching extreme values in cross-attention at late, odd-numbered (user) turns---confirming that the model learns to scrutinize positions where adversarial payloads typically appear.}
	\label{fig:attention_analysis}
\end{figure}

\paragraph{Self-attention patterns.}
As shown in Figure~\ref{fig:attention_aggregate}, safe conversations concentrate \texttt{[CONV\_CLS]} self-attention heavily on early turns---T1--T3 receive approximately $40\%$ of total weight---reflecting predictable topic-setting and cooperative exchange patterns. Jailbreak conversations distribute self-attention more evenly across all turn positions, with later turns (T5--T24) receiving proportionally more weight. Figure~\ref{fig:attention_ratio} quantifies this shift: the JB/Safe self-attention ratio rises from $0.67{\times}$ at T2 to a peak of $2.7{\times}$ at T14, consistent with the model recognizing gradual escalation patterns that span the full conversation rather than relying on any single suspicious turn.

\paragraph{Cross-attention patterns.}
The difference is even more pronounced in cross-attention (right panels of Figures~\ref{fig:attention_aggregate} and~\ref{fig:attention_ratio}). Safe conversations exhibit a dominant spike at T4 ($0.30$ weight, nearly $3{\times}$ any other position), suggesting the model anchors on early assistant responses to confirm benign intent. Jailbreak conversations show more uniform cross-attention with moderate peaks distributed across later turns, corresponding to adversarial payload delivery points. The cross-attention JB/Safe ratio reaches $6{\times}$ at T9, $11{\times}$ at T15, and $16{\times}$ at T21 (Figure~\ref{fig:attention_ratio}, right), with the highest ratios falling consistently on odd-numbered positions (user turns), indicating the model learns to direct token-level scrutiny toward later user turns where harmful requests are most likely to appear in multi-turn escalation attacks.

\paragraph{Architectural implications.}
This attention asymmetry explains the architectural advantage of the hierarchical ConvTransformer: its cross-attention mechanism allows the \texttt{[CONV\_CLS]} token to directly query any turn's encoded representation, enabling detection of late-conversation attacks that fixed-window approaches inherently miss. The increasing JB/Safe ratio---together with its concentration on odd-numbered user turns---further suggests that the model has learned an implicit ``escalation detector'' that up-weights evidence from later user turns proportional to their positional risk, a capability that emerges naturally from training without explicit positional supervision.

\subsection{Ablation Study}

\paragraph{ConvTransformer Depth Ablation}

We vary the number of ConvTransformer decoder layers from $L=1$ to $L=5$ to study the effect of model depth on performance. All variants use the same Multilingual E5-base TurnEncoder and training procedure. As shown in Table~\ref{tab:convtransformer-depth-ablation}, increasing depth from $L=1$ to $L=4$ yields the best overall performance, while deeper models provide diminishing returns.

\begin{table}[!t]
\centering
\caption{ConvTransformer depth ablation on the test set (14{,}038 conversations). All models were trained on 98.6k data. Values are percentages ($\tau=0.5$).}
\label{tab:convtransformer-depth-ablation}
\begin{tabular}{lcccccc}
\hline
\textbf{Layers} & \textbf{AUC-ROC} & \textbf{Acc} & \textbf{F1 Macro} & \textbf{JB Rec} & \textbf{JB Prec} & \textbf{FPR} \\
\hline
$L=1$ & 97.67 & 94.23 & 94.02 & 90.33 & 95.59 & 2.98 \\
$L=2$ & 98.63 & 94.76 & 94.57 & 91.27 & 95.96 & 2.75 \\
$L=3$ & 98.15 & 94.26 & 94.04 & 89.81 & 96.18 & 2.55 \\
$L=4$ & 98.66 & 95.04 & 94.87 & 92.06 & 95.89 & 2.82 \\
$L=5$ & 98.42 & 94.73 & 94.54 & 91.46 & 95.71 & 2.93 \\
\hline
\end{tabular}
\end{table}

$L=4$ achieves the best performance across all primary metrics: highest AUC-ROC (98.66\%), highest accuracy (95.04\%), and highest jailbreak recall (92.06\%). The advantage over $L=1$ is meaningful: +1.7pp JB recall and +0.8pp accuracy, demonstrating that deeper models better capture the multi-step reasoning required for detecting subtle escalation patterns.

$L=3$ shows a slight dip relative to $L=2$ (89.81\% vs.\ 91.27\% recall), suggesting an intermediate-depth instability. $L=5$ marginally underperforms $L=4$, indicating diminishing returns beyond 4 layers for this task and dataset size.

All depths maintain FPR between 2.55\% and 2.98\%. The deeper models' recall gains do not come at the cost of additional false positives.

\begin{figure}[!t]
    \centering
    \begin{subfigure}[t]{0.48\textwidth}
        \centering
        \includegraphics[width=\linewidth, height=5cm]{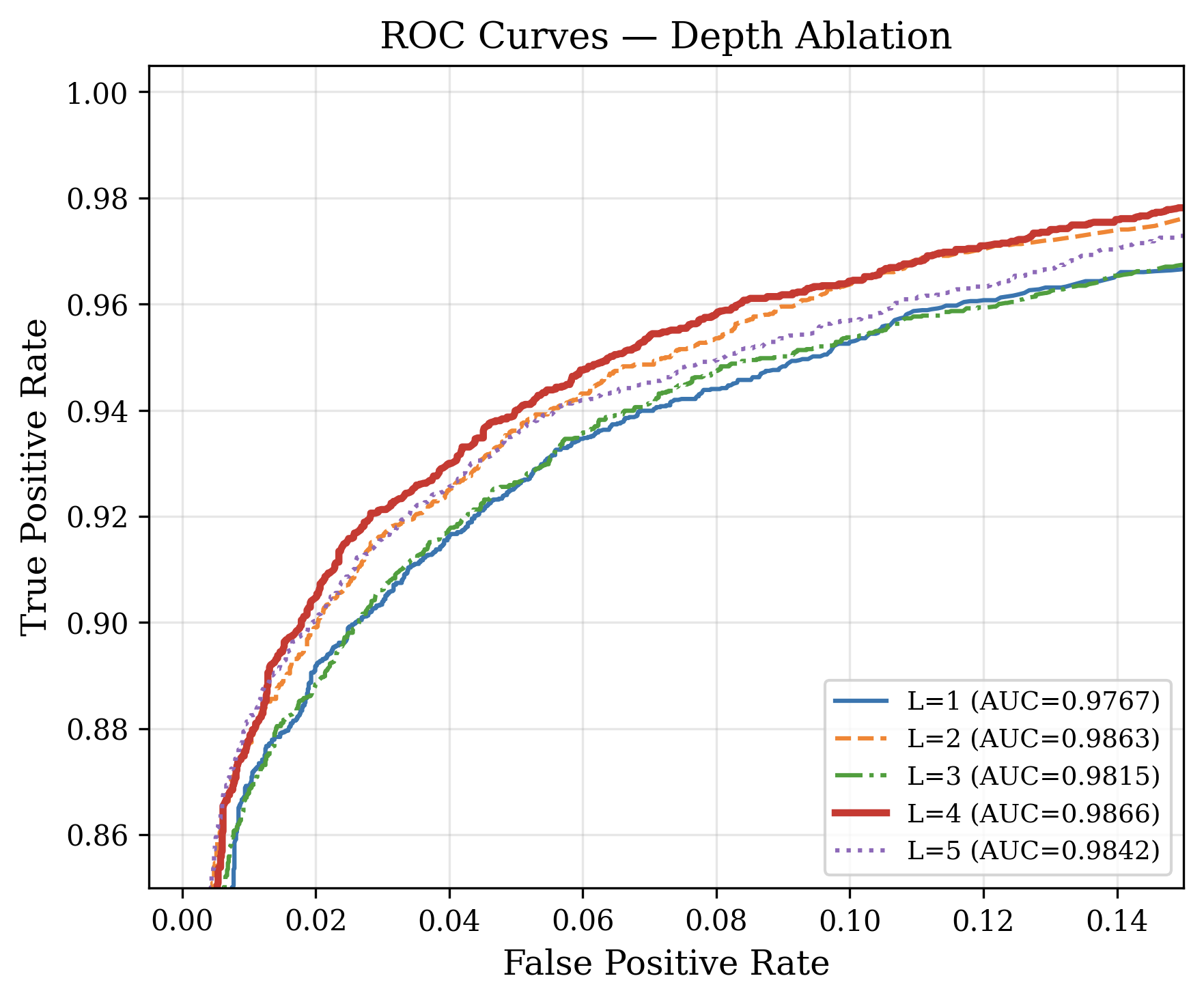}
        \caption{ROC curves by ConvTransformer depth ($L=1$ to $L=5$).}
        \label{fig:roc-depth}
    \end{subfigure}
    \hfill
    \begin{subfigure}[t]{0.48\textwidth}
        \centering
        \includegraphics[width=\linewidth, height=5cm]{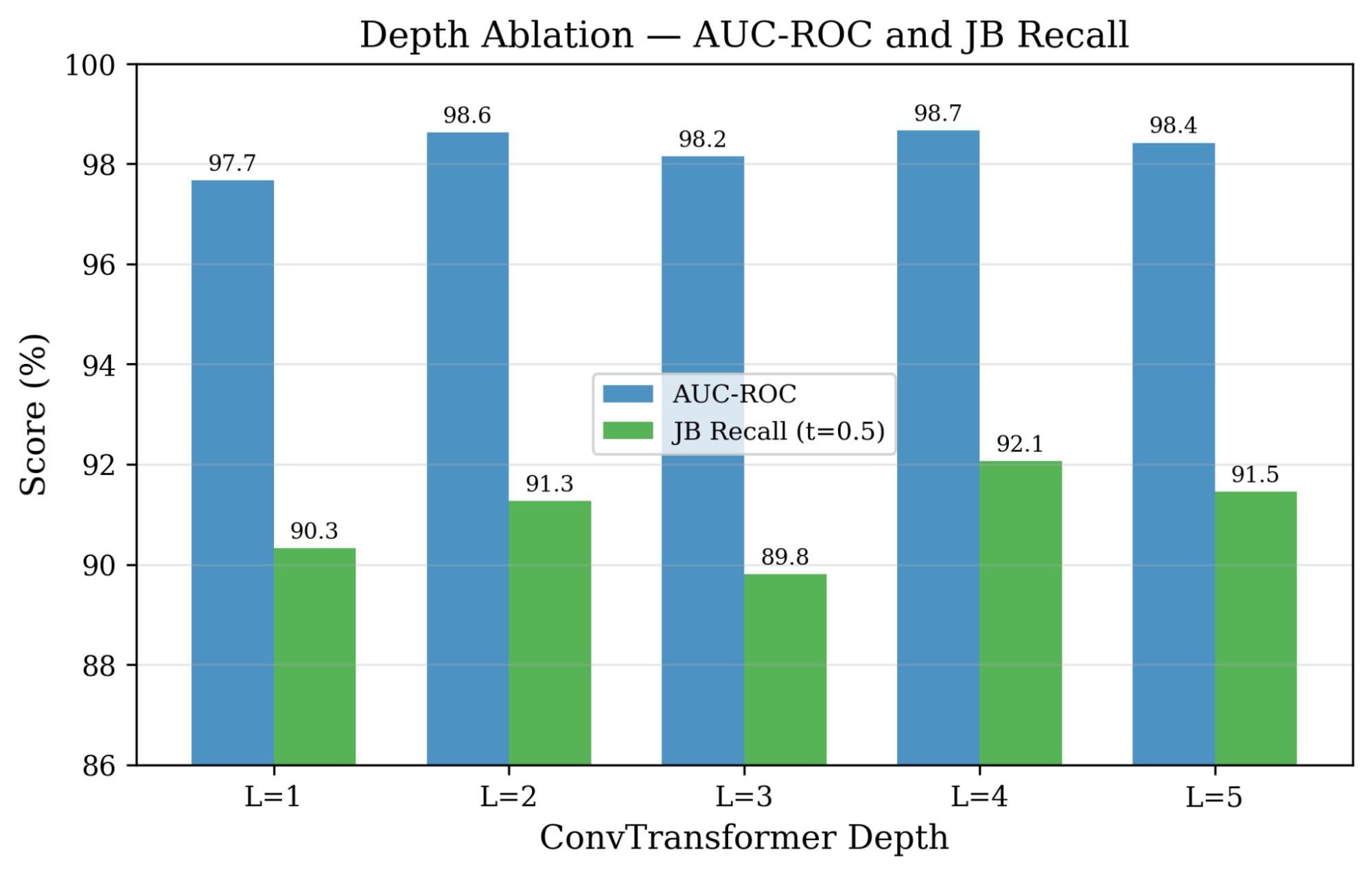}
        \caption{AUC-ROC and JB Recall by ConvTransformer depth ($L=1$ to $L=5$).}
        \label{fig:auc-recall-depth}
    \end{subfigure}
    \caption{Performance trends across ConvTransformer depths.}
    \label{fig:depth-analysis}
\end{figure}

Figure~\ref{fig:depth-analysis} visualizes the impact of model depth on classification quality. 
Figure~\ref{fig:roc-depth} shows that all ConvTransformer variants achieve strong ROC performance, with the deeper models generally tracing curves closer to the upper-left corner, indicating better discrimination. 
Figure~\ref{fig:auc-recall-depth} further highlights the depth-performance trade-off: performance improves from $L=1$ to $L=4$, where both AUC-ROC and jailbreak recall reach their strongest values, while $L=5$ shows a slight decline, suggesting diminishing returns beyond four layers.

\paragraph{Attention Component Ablation}
\label{sec:attention-component-ablation}

\begin{figure}[t]
	\centering
	\includegraphics[width=0.6\linewidth]{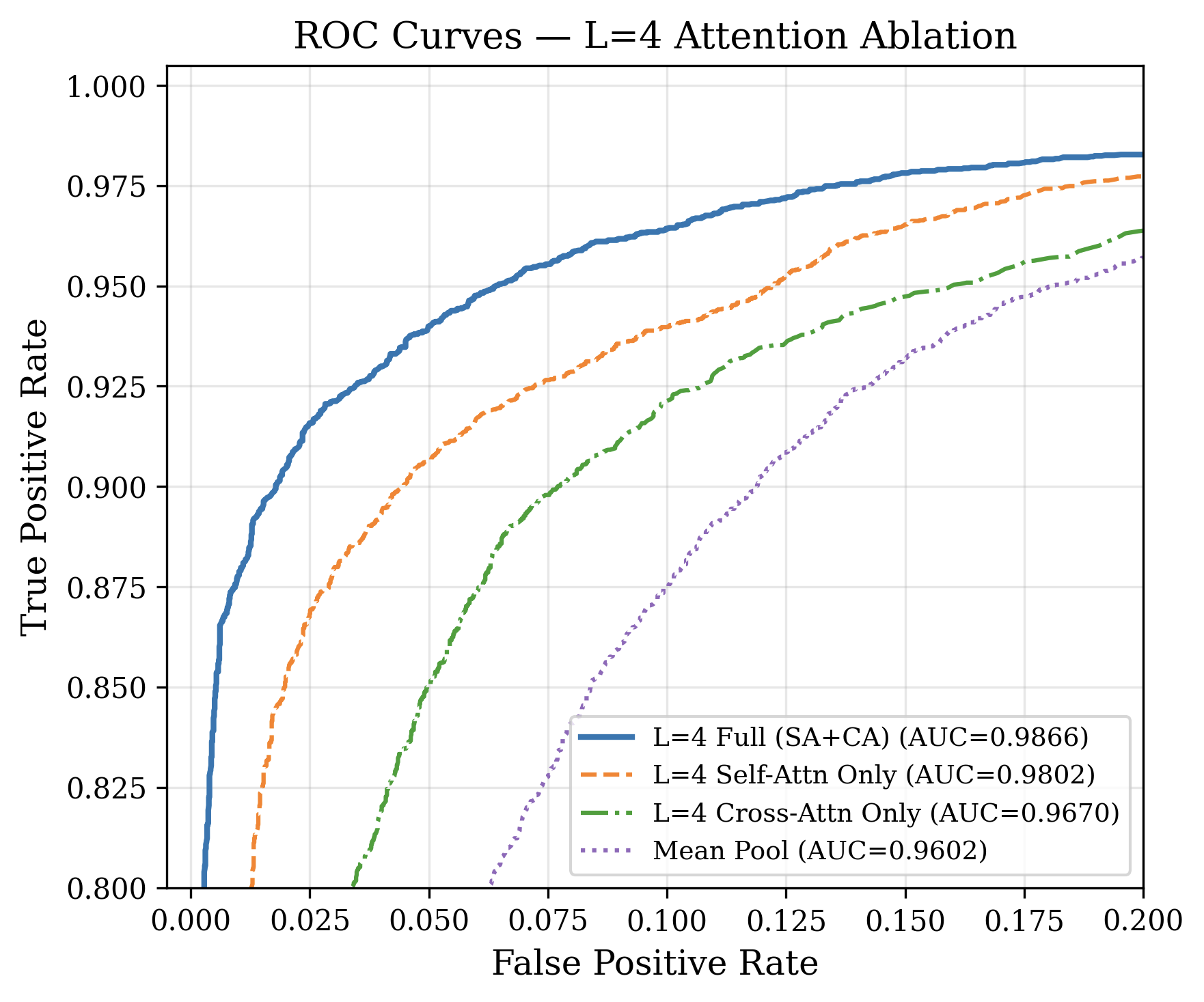}
	\caption{ROC comparison across attention component ablation variants.}
	\label{fig:attention-ablation-roc}
\end{figure}

We ablate the attention components of the $L=4$ ConvTransformer by selectively disabling self-attention or cross-attention. This reveals the distinct contributions of each mechanism at the chosen model depth. Quantitative results are shown in Table~\ref{tab:attention-ablation}, and ROC comparisons are shown in Figure~\ref{fig:attention-ablation-roc}.

At $L=4$ depth, the attention components show strongly differentiated roles. As shown in Table~\ref{tab:attention-ablation}, self-attention captures inter-turn dynamics and achieves competitive recall (90.76\%) but at nearly double the false positive rate (5.08\%). It detects multi-turn escalation patterns by modeling how turns relate to each other, but this sensitivity to structural patterns also triggers on benign instructional conversations.

Cross-attention alone degrades dramatically, reaching 78.30\% recall, a drop of 13.8 percentage points relative to the full model in Table~\ref{tab:attention-ablation}. Without self-attention to provide conversation-level context, cross-attention's token-level content matching cannot recover the distributed signal of gradual escalation attacks. However, it maintains a low FPR (2.96\%), suggesting that its errors are primarily missed detections rather than false alarms.


The full model achieves a synergy that neither component matches alone. Relative to self-attention only, it improves recall by 1.3 percentage points while substantially reducing FPR (2.82\% vs.\ 5.08\%). 

\clearpage 

\begin{table}[t]
	\caption{Attention component ablation on the test set. $L=4$ ConvTransformer, trained on 98.6k data. Values are percentages ($\tau=0.5$).}
	\label{tab:attention-ablation}
	\centering
	\small
	\begin{tabular}{lccccc}
		\toprule
		\textbf{Variant} & \textbf{AUC-ROC} & \textbf{Acc} & \textbf{JB Rec} & \textbf{JB Prec} & \textbf{FPR} \\
		\midrule
		Full (SA+CA)    & 98.66 & 95.04 & 92.06 & 95.89 & 2.82 \\
		Self-Attn Only  & 98.02 & 93.18 & 90.76 & 92.74 & 5.08 \\
		Cross-Attn Only & 96.70 & 89.22 & 78.30 & 94.99 & 2.96 \\
		Mean Pool       & 96.02 & 87.11 & 75.56 & 92.13 & 4.62 \\
		\bottomrule
	\end{tabular}
\end{table}

Relative to cross-attention only, it improves recall by 13.8 percentage points. We also observe that this synergy is substantially stronger at $L=4$ than at $L=1$, where the two components perform more similarly, suggesting that deeper architectures enable richer interactions between the self-attention and cross-attention pathways.

To further isolate the contribution of attention-based conversation-level aggregation, we additionally compare against a non-attentive baseline in which the ConvTransformer cross-turn encoder is replaced with simple mean pooling over turn embeddings, followed by a linear classifier. Both variants share the same Multilingual E5-base TurnEncoder, so any performance difference is attributable to the aggregation mechanism rather than to the turn-level representation. As shown in Table~\ref{tab:attention-ablation}, the full model substantially outperforms this Mean Pool baseline across all metrics, improving AUC-ROC by 2.64 percentage points (98.66\% vs.\ 96.02\%), accuracy by 7.93 points (95.04\% vs.\ 87.11\%), and jailbreak recall by 16.50 points (92.06\% vs.\ 75.56\%), while simultaneously reducing FPR from 4.62\% to 2.82\%. The large recall gap is particularly informative: mean pooling treats all turns equally and dilutes localized adversarial signals, whereas the attention-based ConvTransformer can selectively emphasize the turns and tokens most indicative of escalation. These results indicate that attention-based aggregation, rather than turn-level encoding alone, is the primary driver of the hierarchical model's effectiveness on multi-turn jailbreak detection.

\section{Conclusion}

We introduced a conversation-level jailbreak detector that assigns each dialogue a jailbreak likelihood and enables straightforward deployment via a single decision threshold. Across a challenging out-of-distribution evaluation, our approach consistently outperformed a broad set of baselines, including Claude Opus 4.7, demonstrating strong generalization under domain shift. The proposed hierarchical design avoids expensive long-context concatenation by encoding turns independently and performing lightweight conversation-level reasoning. It also supports selectively attending to fine-grained token evidence when needed, improving detection of multi-turn escalation and reframing behaviors.

Future work will focus on improving robustness to adaptive jailbreak strategies, refining calibration and threshold selection across domains, and further reducing false positives without sacrificing detection coverage.


\bibliographystyle{plainnat}
\bibliography{references} 


\appendix
\section*{Appendix}            
\addcontentsline{toc}{section}{Appendix}  



\section{Evaluation Benchmark Composition}
\label{app:dataset_composition}

Tables~\ref{tab:safe_sources} and~\ref{tab:jb_sources} provide a complete per-source breakdown of the 14{,}038-conversation evaluation benchmark introduced in Section~\ref{sec:datasets}.

\begin{table}[h]
	\centering
	\small
	\caption{Safe conversations in the evaluation benchmark (label $=0$): 8{,}182 total.}
	\label{tab:safe_sources}
	\begin{tabular}{lrlp{5.5cm}l}
		\toprule
		\textbf{Source} & \textbf{N} & \textbf{Turns} & \textbf{Description} & \textbf{Reference} \\
		\midrule
		UltraChat & 2{,}937 & 4--14 & Synthetic open-domain dialogues on science, daily life, technology, and creative topics. & \citep{ding2023ultrachat} \\
		WildChat & 2{,}800 & 4--86 & Real user--ChatGPT interactions with follow-up questions, clarifications, and diverse intents. & \citep{zhao2024wildchat} \\
		ShareGPT & 666 & 4--72 & Community-shared ChatGPT conversations spanning coding help, writing, and general Q\&A. & \citep{sharegpt2023} \\
		OASST2 & 262 & 4--6 & Crowd-sourced assistant dialogues covering factual questions and task completion. & \citep{kopf2023openassistant} \\
		Synthetic hard neg. & 388 & 4--58 & LLM-generated benign conversations that structurally resemble attacks but contain no adversarial intent. & Generated \\
		Long hard neg. & 278 & 4--102 & Extended ($20+$-turn) benign conversations; controls for length bias in the detector. & Generated \\
		Benign follow-ups & 390 & 4 & Short benign conversations with polite user follow-ups; targets gratitude patterns confused with post-refusal jailbreak turns. & Generated \\
		SafeMTData (safe) & 192 & 4--12 & Multi-turn conversations on sensitive-adjacent topics (e.g., DNS, cybersecurity) with genuinely benign intent. & \citep{ren2024safemtdata} \\
		Security discussions & 190 & 4--30 & Benign conversations on penetration testing and red-teaming methodology. & Generated \\
		Other & 79 & 4--5 & Empathetic dialogues and safe adversarial-ML research conversations. & Mixed \\
		\bottomrule
	\end{tabular}
\end{table}

\begin{table}[h]
	\centering
	\small
	\caption{Jailbreak conversations in the evaluation benchmark (label $=1$): 5{,}856 total.}
	\label{tab:jb_sources}
	\begin{tabular}{lrlp{5.5cm}l}
		\toprule
		\textbf{Source} & \textbf{N} & \textbf{Turns} & \textbf{Description} & \textbf{Reference} \\
		\midrule
		XGuard & 1{,}003 & 4--14 & Red-team attacks using persona adoption and role-play framing. & \citep{xguard2025} \\
		Red Queen & 802 & 5--7 & Short targeted manipulation exploiting trust or authority claims. & \citep{jiang2025redqueen} \\
		SafeDialBench & 408 & 6--16 & Multi-turn safety dialogues testing gradual boundary-pushing across harm categories. & \citep{cao2025safedialbench} \\
		carl213 & 111 & 4--26 & Cipher-encoded attacks using user-defined word-substitution mappings. & \citep{broomfield2024multiturn} \\
		Duke Safety & 75 & 5 & Short gradual-escalation attacks across 37 harm categories. & \citep{chen2025maliciouseducator} \\
		Long escalation & 828 & 4--32 & Extended ($15+$-turn) conversations incrementally steered toward harmful requests. & Generated \\
		Synth. escalation & 465 & 9--20 & LLM-generated gradual-escalation attacks with controlled escalation speed. & Generated \\
		FRACTURED-SORRY & 94 & 7--15 & Multi-step attacks with benign intermediary turns fragmenting harmful intent. & \citep{priyanshu2024fracturedsorry} \\
		Synth. sudden & 461 & 9--21 & Benign conversations with a single adversarial turn inserted at a random late position. & Generated \\
		Long sudden & 268 & 19--31 & Long benign conversations with adversarial insertion; tests detection under signal dilution. & Generated \\
		Ctx. narrowing & 469 & 7--15 & Benign multi-turn prefix followed by a final user turn demanding harmful reformatting of the context. & Generated \\
		Prompt injection & 318 & 2--12 & Short instruction-override attacks (``ignore previous instructions'', system-prompt extraction, hypothetical framing). & Generated \\
		LAD & 152 & 4--18 & Gradual escalation in technical domains (SCADA, AVs, network security) from benign inquiry to exploit requests. & \citep{kulkarni2024lad} \\
		Gradual technical & 402 & 4--30 & Benign technical questions (cybersecurity, programming) escalating into explicit exploit/attack-code requests. & Generated \\
		\bottomrule
	\end{tabular}
\end{table}

\end{document}